\documentclass[runningheads]{llncs}
\usepackage[T1]{fontenc}
\usepackage{graphicx}
\usepackage{xcolor}
\usepackage[colorlinks,citecolor=green]{hyperref}
\usepackage{algorithmic}
\usepackage[ruled, lined, longend, linesnumbered]{algorithm2e}
\usepackage[subtle,margins=normal,leading=normal]{savetrees}
\begin{document}
\title{Warm Start Active Learning with Proxy Labels \& Selection via Semi-Supervised Fine-Tuning}
\titlerunning{Proxy Semi-Supervised Active Learning}
%
\author{Vishwesh Nath\inst{} \and
Dong Yang\inst{} \and
Holger R. Roth\inst{} \and
Daguang Xu\inst{}}
%
\authorrunning{V. Nath et al.}
%
\institute{NVIDIA
\email{}\\
\url{}
\email{vnath@nvidia.com}}
\maketitle              
\begin{abstract}
Which volume to annotate next is a challenging problem in building medical imaging datasets for deep learning. One of the promising methods to approach this question is active learning (AL). However, AL has been a hard nut to crack in terms of which AL algorithm and acquisition functions are most useful for which datasets. Also, the problem is exacerbated with which volumes to label first when there is zero labeled data to start with. This is known as the cold start problem in AL. We propose two novel strategies for AL specifically for 3D image segmentation. First, we tackle the cold start problem by proposing a proxy task and then utilizing uncertainty generated from the proxy task to rank the unlabeled data to be annotated. Second, we craft a two-stage learning framework for each active iteration where the unlabeled data is also used in the second stage as a semi-supervised fine-tuning strategy. We show the promise of our approach on two well-known large public datasets from medical segmentation decathlon. The results indicate that the initial selection of data and semi-supervised framework both showed significant improvement for several AL strategies.

\end{abstract}
\section{Introduction}

Active learning (AL) \cite{settles2012active} for medical image segmentation can help reduce the burden of annotation effort as it focuses on selection of the \textit{more} relevant data that may lead to a better performing model. However, there is disagreement among which acquisition function for ranking the unlabeled data is the best \cite{sener2017active,bengar2021deep,bengar2021reducing,simeoni2021rethinking,chitta2018large} and also which framework is the best for uncertainty generation \cite{nguyen2021measure,wang2018test,gal2017deep,nath2021power,beluch2018power,nath2020diminishing,gal2016dropout}. Apart from these challenges for AL, there are two other critical problems to be addressed as well: (1) When faced with an entirely unlabeled pool of data, how does one select the initial set of data to start annotation? This falls under the problem of the ``cold start'' in AL \cite{yuan2020cold,houlsby2014cold}. (2) Current acquisition functions utilize supervised models which learn using labeled data. There is a lack of 3D-based AL algorithm that leverages the knowledge of unlabeled data in uncertainty generation and data ranking. The models based on semi-supervised learning (SSL) can become better data selectors \cite{simeoni2021rethinking,wang2021annotation}. So far SSL has only been shown to be effective for helping AL algorithms in generic computer vision tasks \cite{simeoni2021rethinking,bengar2021reducing,bengar2021deep} or 2D medical segmentation \cite{wang2021annotation}. Unfortunately, SSL models for 3D segmentation are not straightforward, as the entire 3D volume cannot be used as input, unlike the 2D variants due to memory constraints. They face the challenge of appropriate patch selection, which needs to focus on the desired region of interest (ROI). In this work, we propose novel strategies to tackle these challenges. The experiments have been performed on two well-known large public computed tomography (CT) datasets \cite{simpson2019large}.

\section{Related Work}

\textbf{Cold Start AL:} The knowledge of which data to annotate first, given a completely unlabeled pool of data, can serve as an excellent starting point for the active learning process. Prior work has been done for natural language processing \cite{ash2019deep,yuan2020cold}. In \cite{yuan2020cold} an embedding is generated for sentences using a pre-trained transformer model. Clustering is performed on the embeddings; the sentences closest to cluster centers are selected first for annotation. Compared to random selection, such a strategy improves the data annotation process when there is zero labeled data. This is highly relevant in the medical imaging domain where there is an abundance of unlabeled data, but annotations are costly and often unavailable.

\noindent \textbf{Semi-Supervised AL:} There is prior work that has shown that semi-supervised learning is beneficial for AL where it makes the models a better selector of annotation data while improving the performance \cite{bengar2021reducing,simeoni2021rethinking}. However, these have only been shown for either classification \cite{simeoni2021rethinking} or for 2D segmentation \cite{wang2020semi,lai2021joint}. There is no AL work combined with 3D semi-supervised learning to the best of our knowledge.

Meanwhile, multiple methods have been proposed that fall under semi-supervised learning techniques for 3D segmentation, such as \cite{xia20203d} which utilizes an ensemble of models for SSL, \cite{yu2019uncertainty} which uses a student-teacher method, \cite{li2020shape} which proposes to use a shape constraint for SSL, as well as prediction consistency-based approaches or contrastive approaches \cite{tang2021self}. Ensembles are computationally expensive, while student-teacher and shape constraints are not ideal for AL either. Hence, we explore towards consistency-based approaches for AL SSL.

\begin{figure}[!htb]
\begin{center}
    \includegraphics[width=\linewidth]{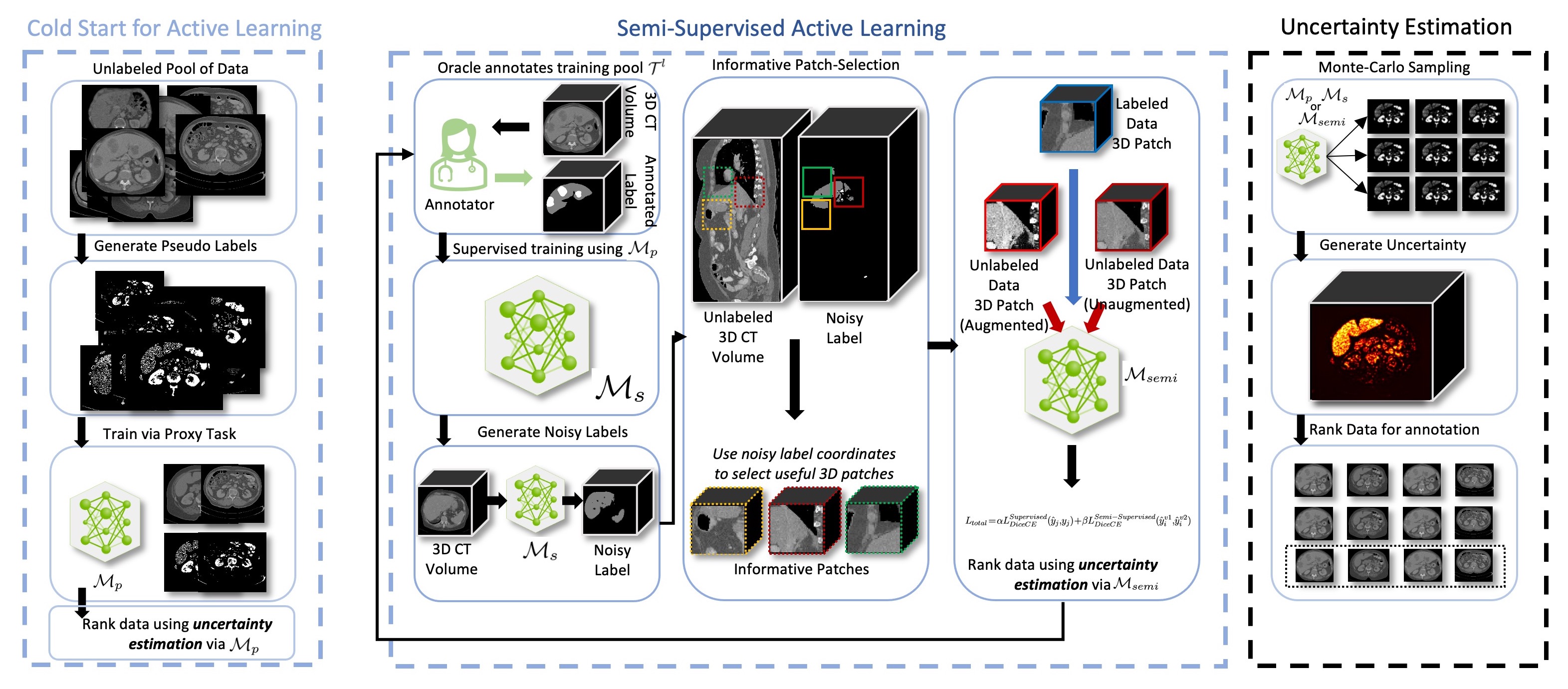}
    \caption{The proposed pipeline of the entire AL framework. Left to right: First, we generate the pseudo-labels for a proxy task to perform data ranking via uncertainty. The pre-trained model and the ranking both are used in the semi-supervised AL framework, which is a two-stage learning process. After semi-supervised learning is done, the final model is used to select data for annotation to repeat the cyclic process.}
    \label{fig1:overview}
\end{center}
\end{figure}

\section{Proposed Method}

The proposed method starts with ranking of a zero-labeled data pool via a proxy task that utilizes pseudo labels, which can also be considered as a self-supervised learning task (Ref Fig. \ref{fig1:overview}). After the annotation of selected unlabeled data, a fully supervised training is performed. The trained model is consecutively fine-tuned via a semi-supervised approach. Finally, the fine-tuned semi-supervised model is used for data selection of which 3D volumes should be labeled next by the annotator. The cycle continues till the desired performance is achieved. Please refer to Algorithm. \textcolor{red}{1} for more details.

We utilize a U-Net \cite{ronneberger2015u} like 5-level encoder-decoder network architecture with residual blocks at every level. In the experiments, only the number of output channels are modified based upon the number of classes of the segmentation task. We consistently utilize a softmax function for activation to probability maps $p(x_i)$, for proxy, supervised and semi-supervised learning.

\subsection{Pre-Ranking of Data via Pseudo Labels as a Proxy Task} \label{subsection:proxy}

Consider an unlabeled pool of data $\mathcal{U}$ which consists of $n$ samples, forming a set as $\mathcal{U}=\{x_1, x_2, ... x_n\}$. To generate a pseudo label denoted as $y^p_i$ for a given sample data point $x_i$, we threshold $x_i$ within a typical abdominal soft-tissue windows for CT data (W:50 L:125) Hounsfield Units (HU)\footnote{\url{https://radiopaedia.org/articles/windowing-ct}}, then utilize largest connected component analysis to select ROIs that are well connected. Intuitively, the expectation is to select all major organs of the abdomen as foreground. All prior approaches regarding pseudo-labels have been 2D or on a slice-by-slice basis~\cite{ouyang2020self,wang2021annotation}. Here, we provide an efficient and effective 3D extension. 

After the generation of pseudo labels, we form a pseudo-labeled dataset $\mathcal{U}^p=\{(x_1, y_1^p), (x_2, y_2^p), ... (x_n, y_n^p)\}$. A proxy model is trained using the pseudo-labels that utilizes a combination of Dice and Cross-entropy (DiceCE) loss denoted as $L_{DiceCE}$ as outlined in \cite{isensee2021nnu}. Please note that this is a binary segmentation task.

The proxy-trained model tackles the cold start AL problem with two specific advantages. First, it acts as a pre-trained model, which is a good initialization compared to random initialization. Since the model is already trained on a pseudo segmentation task, it is better suited to prior initialization as compared to random initialization. Second, it allows for data uncertainty estimation of all the unlabeled volumes $\mathcal{U}$ that it was trained on. Monte Carlo simulation is used to generate multiple predictions with dropout enabled during inference to estimate the uncertainty \cite{gal2016dropout}. More details on uncertainty estimation can be found in section \ref{subsection:uncertainty}. Uncertainty was generated for all unlabeled data via the proxy model $\mathcal{M}_p$, thereafter the data was sorted for selection based on uncertainty generated by $\mathcal{M}_p$ per data point. The most uncertain volumes are selected for the initial pool denoted as $\mathcal{T}^l$ with the pre-trained checkpoint of the model. The initially selected training pool data points are annotated by the annotator for supervised learning. Once a volume is labeled it is removed from $\mathcal{U}$. $k$ volumes are selected for $\mathcal{T}^l$. At every active learning iteration, $k$ volumes are added to $\mathcal{T}^l$.

\subsection{Fully Supervised Training}

The fully supervised model denoted as $\mathcal{M}_{s}$ is trained on all the available labeled data from the training pool $\mathcal{T}^l$. $L_{DiceCE}$ as defined in section \ref{subsection:proxy} is used for training. The network uses all the layers of the pre-trained checkpoint from the proxy task. If the number of classes for segmentation are more than 2, then the number of output channels are increased and the last layer of the proxy model $\mathcal{M}_p$ is ignored when loading the pre-trained model.

\subsection{Semi-Supervised Training}

Due to the computational expense of 3D training and maintaining a balance between labeled and unlabeled data, we cannot use all the unlabeled volumes for training. Hence, $\mathcal{M}_{s}$ is used to select the most \textit{certain} 3D volumes from the remaining unlabeled pool of data $\mathcal{U}$. The number of selected unlabeled volumes for semi-supervised training is equivalent to the number of labeled data in $\mathcal{T}^l$. The biggest challenge with 3D training of unlabeled data is the selection of the more informative cubic patches that contain the specific ROI, which is crucial for the segmentation task. To overcome this challenge, we generate noisy labels denoted as $\hat{y}^n$ of the selected most certain unlabeled 3D volumes using $\mathcal{M}_{s}$. The generated labels are so called ``noisy'' because the $\mathcal{M}_s$ is trained on a small sample size and hence is a low-performance model which is unlikely to lead to good and clean predictions. Thus, we use a threshold $\tau$ on the probability maps $p(x_i)$ to remove false positive predictions.

The semi-supervised learning is performed as a fine-tuning step to the existing trained model $\mathcal{M}_s$. The model trained from the semi-supervised learning framework is denoted as $\mathcal{M}_{semi}$. The training process utilizes both labeled $\mathcal{T}^l$ and unlabeled data $\mathcal{T}^n$. Please note that all samples from $\mathcal{T}^l$ are also used in semi-supervised training (Fig. \ref{fig1:overview}). The training process for a single step is defined by the following loss function:
\begin{equation}
    L_{total} = \alpha L_{DiceCE}^{Supervised}(\hat{y}_j, y_j)  +  \beta L_{DiceCE}^{Semi-Supervised}(\hat{y}^{v1}_i, \hat{y}^{v2}_i)
\end{equation}
where supervised and semi-supervised, $L_{DiceCE}$ loss is the same as in section \ref{subsection:proxy}. The semi-supervised loss is based on a randomly chosen unlabeled volume $x^U_i$ from which a cubic patch is selected that focuses on the required ROI. This ROI patch is extracted based on the noisy label $\hat{y}^n_i$. Consecutively, two views $x^{v1}_i$, $x^{v2}_i$ of the cubic patch are generated, $x^{v1}_i$ is kept unaugmented for best prediction from $\mathcal{M}_{semi}$ while the other view $x^{v2}_i$ is augmented. $\hat{y}^{v1}_i$, $\hat{y}^{v2}_i$ are predictions from $\mathcal{M}_{semi}$. $\alpha$ and $\beta$ allow for weighing the loss function terms for supervised and semi-supervised, respectively.  

\begin{algorithm}[!t]
\scriptsize
\caption{Semi-Supervised AL Algorithm}
\textbf{Input:} Unlabeled Pool $\mathcal{U}$\\
Generate Pseudo Labels for $\mathcal{U}$ forming $\mathcal{U}^p=\{(x_1, y_1^p), ... (x_n, y_n^p)\}$ \\
Train $\mathcal{M}_p$ using $\mathcal{U}^p$, generate uncertainty for $\mathcal{U}$ \\
Sort $\mathcal{U}$ based on uncertainty via $\mathcal{M}_p$\\ 
Select  most uncertain $k$ data from $\mathcal{U}$ for annotation, add $k$ to $\mathcal{T}^l_j$ \\
\For{$j \in \{0,1 ...\}$}{
$\mathcal{M}_s$ = Supervised Training on $\mathcal{M}_p$ using $\mathcal{T}^l_j$ \\
\eIf{Semi-Supervised is True}{
Extract most certain \textit{len($\mathcal{T}^l_j$)}data from $\mathcal{U}$ \& add to $\mathcal{U}_n$ \\
Generate Noisy Labels for $\mathcal{U}_n$ \& form $\mathcal{T}^n_j$ \\
$\mathcal{M}_{semi}$ = Semi-Supervised Fine-Tuning on $\mathcal{M}_s$ using $\{ \mathcal{T}^l_j ,\mathcal{T}^n_j\}$ \\
Generate uncertainty via $\mathcal{M}_{semi}$ for $\mathcal{U}$ \\
Select the most uncertain $k$ data from $\mathcal{U}$ \& add to $\mathcal{T}^l_j$ \\
}{
Generate uncertainty via $\mathcal{M}_{s}$ for $\mathcal{U}$ \\
Select the most uncertain $k$ data from $\mathcal{U}$ for annotation \& add to $\mathcal{T}^l_j$}{}
}

\label{alg:only_one}
\end{algorithm}

\subsection{Uncertainty Estimation \& AL} \label{subsection:uncertainty}

We use epistemic or model uncertainty as defined in \cite{chitta2018large,yang2017suggestive} for variance and entropy. The uncertainty measure is estimated based on the number of MC simulations $m$. To obtain a score, we use the mean across all the 3D voxels in the 3D uncertainty map $\mathcal{H}$.


\section{Datasets \& Experiments}

\subsection{Datasets} \label{subsection:datasets}

\textbf{Liver \& Tumor:} This dataset comes from task 3 of Medical Segmentation Decathlon (MSD) \cite{simpson2019large}. It consists of 131 3D CT volumes. The data was split into 105 volumes for training and 26 for validation. The 105 volumes represent the unlabeled pool $\mathcal{U}$ for experimentation. The validation set was kept consistent for all experiments ranging from pseudo-label pre-training to AL results. For pre-processing, all data were re-sampled to $1.5\times1.5\times2$ $mm^3$ spacing for rapid experimentation. Augmentation of shifting, scaling intensity and gaussian noise were used. The CT window was used as per \cite{he2021dints}.

\noindent
\textbf{Hepatic Vessels \& Tumor:} This dataset comes from task 8 of MSD and consists of 303 3D CT volumes. The data was split into 242 volumes for training and 61 for validation. The 242 volumes represent $\mathcal{U}$ for experimentation. The validation set was kept consistent for all experiments ranging from pseudo-label pre-training to AL results. Pre-processing is similar as for liver \& tumor except for CT window for normalization which was adapted as per \cite{he2021dints}.

\subsection{Experiments}

\begin{figure}[!htb]
\begin{center}
    \includegraphics[width=\linewidth]{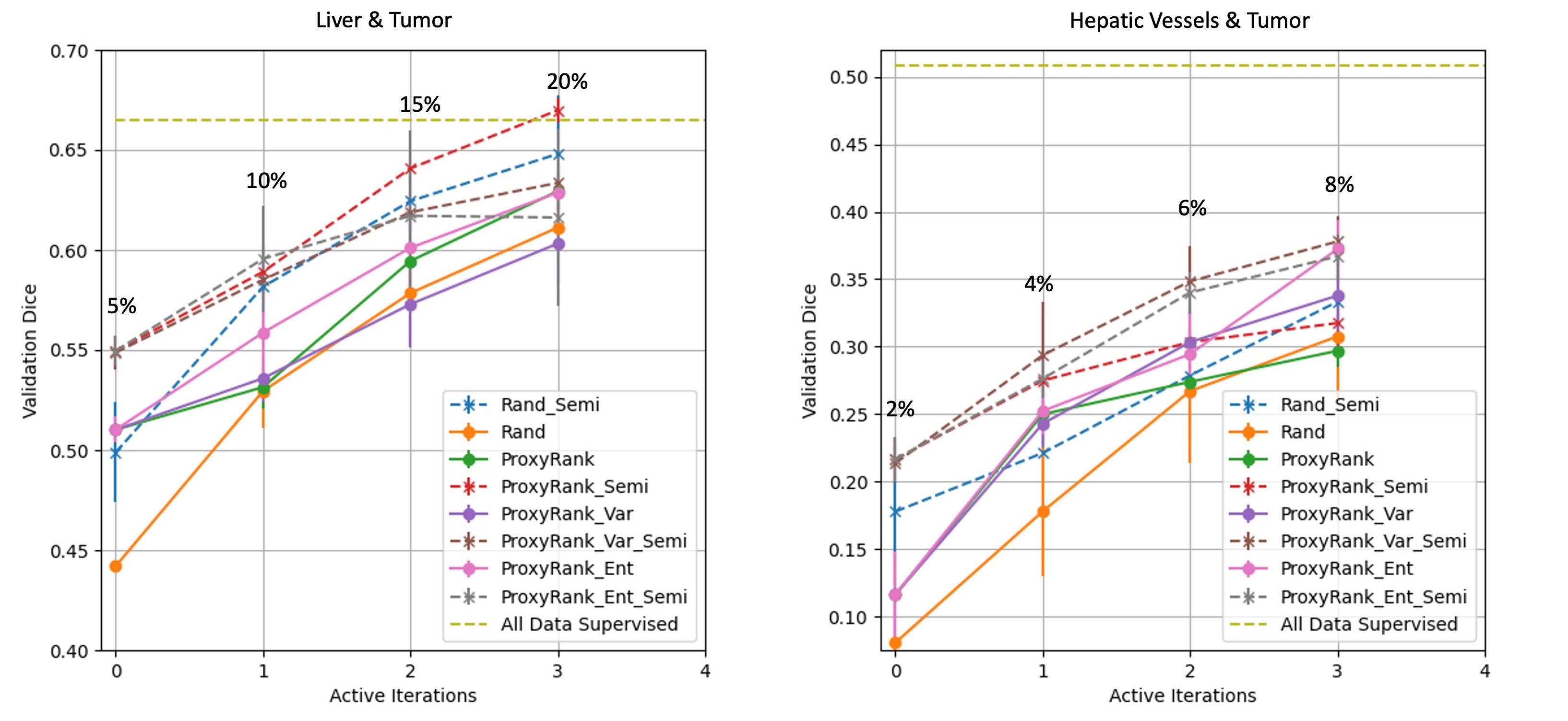}
    \caption{The percentage represents the amount of data used at the current active iteration with respect to all available data. Solid lines represent fully supervised approaches and dashed lines represent semi-supervised approaches. Solid lines are baselines, while dashed lines include our proposed methods. ``Proxyrank'' indicates that the initial training pool was estimated using proxy task. ``Var'' $\rightarrow$ ``Variance'' \cite{yang2017suggestive}, ``Ent'' $\rightarrow$ ``Entropy'' \cite{nath2020diminishing}, ``Rand'' $\rightarrow$ ``Random'', ``Semi'' $\rightarrow$ ``Semi-supervised learning''. Refer Supplementary Table 1. for more details.}
    \label{fig2:al_results}
\end{center}
\end{figure}

\textbf{Proxy Training Hyper-parameters:} All proxy models trained with pseudo-labels were trained for 100 epochs on all data; the validation set for selecting the best model is consistent with the other experiments. The validation frequency was once every 5 epochs. $L_{DiceCE}$ loss was used for training. Cubic patch size of $96\times96\times96$ was used with random selection of patches from the 3D CT volumes. Learning rate was set to $1e^{-4}$ and Adam optimizer \cite{kingma2014adam} was used.

\noindent
\textbf{Uncertainty Hyper-parameters:} To estimate uncertainty, $m$ was set to 10 and the dropout ratio was set to 0.2. A dropout layer was kept at the end of every level of the U-Net for both the encoder and decoder.

\begin{figure}[!t]
\begin{center}
    \includegraphics[width=\linewidth]{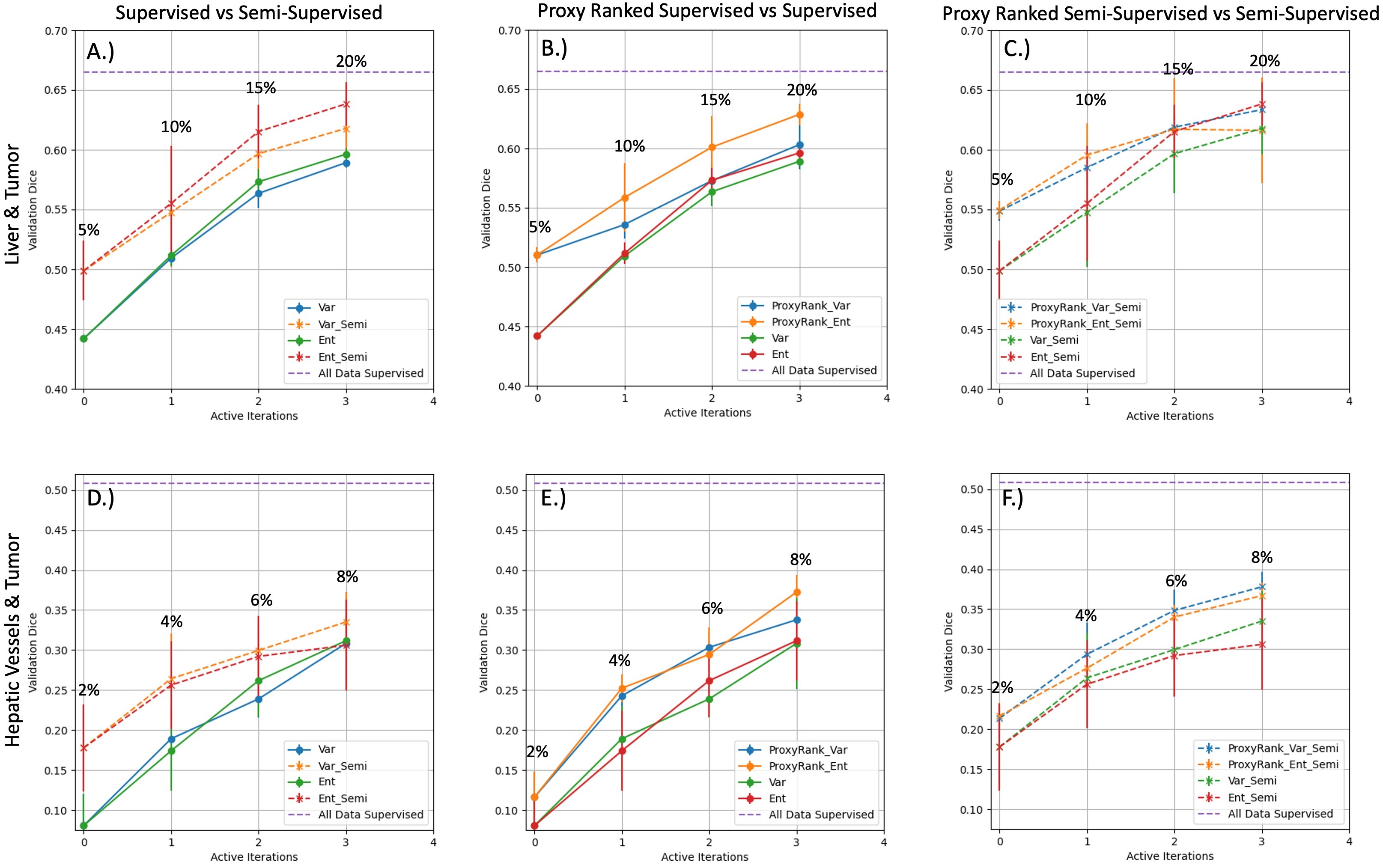}
    \caption{Top row: results for hepatic vessels \& tumor. Bottom row: results for liver \& tumor. The percentage represents the amount of data used at current active iteration with respect to all available data. Solid lines are baselines, while dashed lines include our proposed methods. ``Proxyrank'' indicates that the initial training pool was estimated using proxy task. ``Var'' $\rightarrow$ ``Variance'' \cite{yang2017suggestive}, ``Ent'' $\rightarrow$ ``Entropy'' \cite{nath2020diminishing}, ``Rand'' $\rightarrow$ ``Random'', ``Semi'' $\rightarrow$ ``Semi-supervised learning''. Refer Supplementary Table 1. for more details.}
    \label{fig3:proxy_results}
\end{center}
\end{figure}

\noindent
\textbf{AL Acquisition Functions:} To gauge the benefit of the combined strategies of semi-supervised learning with proxy ranking, we compare them with random based acquisition and also with two well-known AL acquisition functions of entropy and variance \cite{gal2016dropout,chitta2018large,yang2017suggestive}. We also utilize proxy ranking by itself as a strategy for acquisition of data, as that eliminates the need to compute uncertainty at the end of every active iteration. Advanced acquisition functions such as coreset \cite{sener2017active}, maximum-cover \cite{yang2017suggestive}, knapsack \cite{kuo2018cost}, Fisher information \cite{sourati2018active} were not tested as the purpose of this study was to evaluate if the proposed strategies are beneficial for basic components of AL. Though more advanced acquisition functions could be added.

\noindent
\textbf{SSL AL with Proxy Ranking:} One active iteration consists of supervised learning with the existing labeled data and a secondary semi-supervised fine-tuning if needed. All AL experiments were executed for $j=4$ active iterations. For both datasets, liver \& tumor and hepatic vessels \& tumor, $k$ was set to 5 volumes which were being acquired at every active iteration to be annotated by the annotator. The validation set was kept the same for all active iterations as also defined in section \ref{subsection:datasets} for all datasets. 

\begin{figure}[!t]
\begin{center}
    \includegraphics[width=\linewidth]{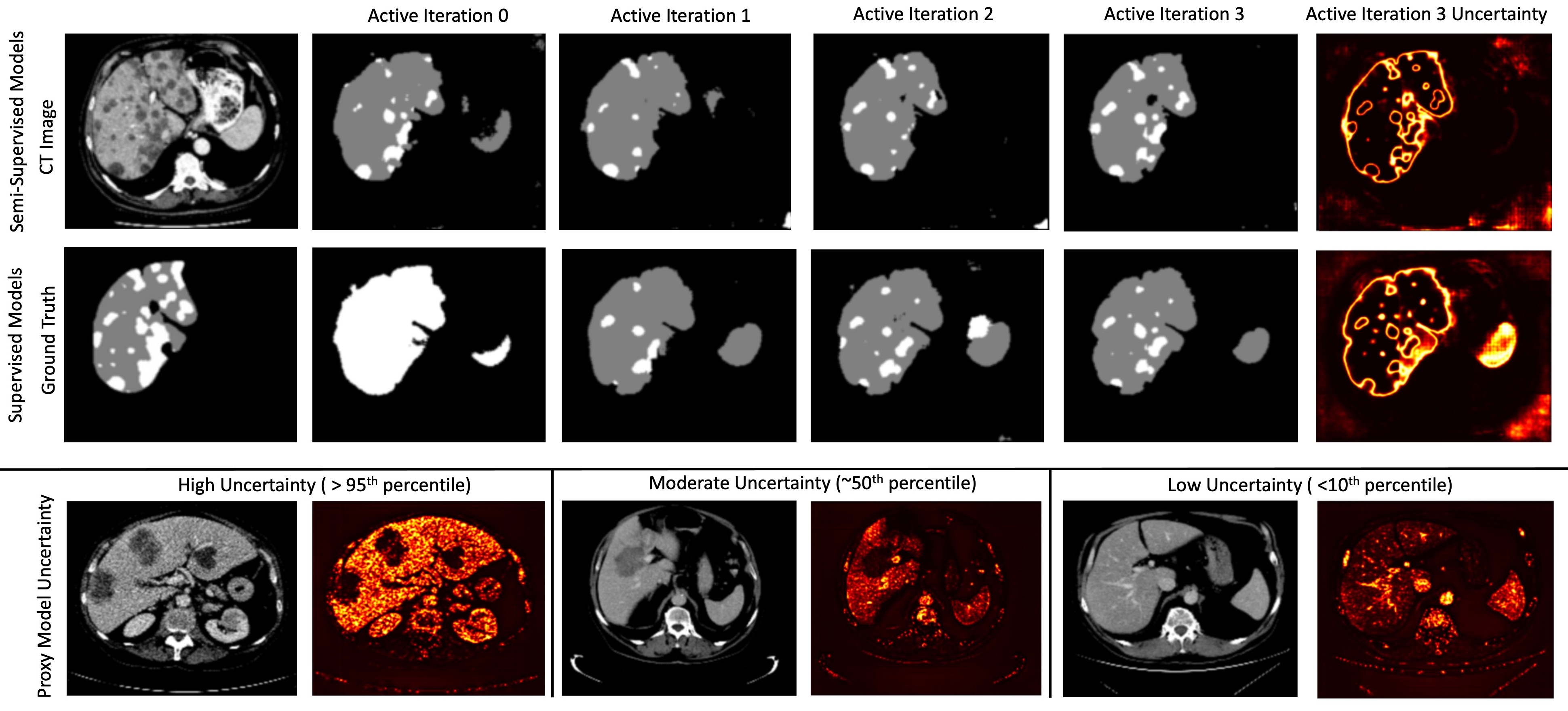}
    \caption{Predictions from successive active iterations. Top row: Supervised models, middle row: Semi-supervised models. The last column shows uncertainty based on the last active iteration. Bottom row: uncertainty on data using proxy task model ranging from high to low}
    \label{fig4:qual_results}
\end{center}
\end{figure}

\begin{table*}[!b]
  \centering
  \scriptsize
  \begin{tabular}{p{2.5cm} p{2.0cm} p{2.0cm} p{2.0cm} p{2.0cm}  }
     \hline
     \multicolumn{5}{c}{\textbf{Proxy Rank \& Semi-Supervised Learning Adv.}} \\
     \hline
     \multicolumn{5}{c}{\textbf{Supervised}} \\
     \hline
     \textit{Dataset} & Variance & Entropy & Proxy Variance & Proxy Entropy  \\\hline
     Liver \& Tumor & $0.5891 \pm 0.0039$\ & $0.5962 \pm 0.0048$ & $0.6032 \pm 0.0207$\ & $0.6287 \pm 0.0092$ \\
     Hepatic \& Tumor & $0.3083 \pm 0.0568$\ & $0.3116 \pm 0.0494$ & $0.3378 \pm 0.0257$\ & $0.3723 \pm 0.0217$\\
     \hline
     \multicolumn{5}{c}{\textbf{Semi-Supervised}} \\
     \hline
     \textit{Dataset} & Variance & Entropy & Proxy Variance & Proxy Entropy  \\\hline
     Liver \& Tumor & $0.6178 \pm 0.0218$\ & $0.6335 \pm 0.0183$ & $\textbf{0.6383} \pm 0.0091$\ & $0.6162 \pm 0.0439$ \\
     Hepatic \& Tumor & $0.3351 \pm 0.0379$\ & $0.3060 \pm 0.0567$ & $\textbf{0.3779} \pm 0.0188$\ & $0.3670 \pm 0.0162$ \\
     \hline
  \end{tabular}
  \caption{The validation Dice at the end of last active iteration are reported to show the benefits of proxy ranking and semi-supervised learning}
  \label{tab:method_select}
\end{table*}

We utilize a fixed number of steps strategy for an equivalent amount of training per active iteration instead of a fixed number of epochs. 2,000 steps were used for both datasets. These steps were estimated from all data training. Learning rate, optimizer and patch size were kept consistent with proxy task hyper-parameters. The selected patches had a foreground and background ratio of 1:1. For semi-supervised learning with noisy labels, this ratio was kept at 9:1. The semi-supervised loss function hyper-parameters $\alpha$ and $\beta$ were set to 1 and 0.001, respectively. The threshold $\tau$ for probability maps $p(x_i)$ was set to 0.9 for noisy label generation.

All AL strategies were repeated three times with different random initializations. The mean and standard deviation of Dice scores are reported in plots and tables. 

\noindent
\textbf{Adv. of Proxy Ranking \& SSL:} To gauge the advantage of proxy ranking for the unlabeled data, we conduct experiments with and without proxy rank to assess both supervised and semi-supervised training. We also gauge the advantage of using semi-supervised over supervised learning. All AL settings are kept the same as described above.

\noindent
\textbf{Implementation:} All deep learning networks were implemented using PyTorch v1.10. MONAI v0.8 \footnote{\url{https://github.com/Project-MONAI/MONAI}} was used for data transformations and pre-processing. The experiments were conducted on Tesla V100 16GB \& 32GB GPUs.

\section{Results}


\textbf{SSL AL with Proxy Ranking:} Quantitatively, semi-supervised \& proxy-ranked approaches provide a benefit over just using supervised methods for AL (Ref Fig. \ref{fig2:al_results}). It should also be noted that the proxy ranking of data is helpful at the first active iteration for both datasets and gives a better starting point for the AL models for both supervised and also semi-supervised methods. Surprisingly, the best performing method for liver \& tumor is given just proxy ranking coupled with semi-supervised learning. While for hepatic vessel \& tumor the best performance is given by the variance acquisition function coupled with semi-supervised learning.


\noindent \textbf{Adv. of Proxy Ranking \& SSL:} Quantitatively, the semi-supervised approaches offer a significant benefit for both datasets (Fig. \ref{fig3:proxy_results}A \& \ref{fig3:proxy_results}D). A similar observation can be made when the proxy ranking is added to either supervised (Fig. \ref{fig3:proxy_results}B \& \ref{fig3:proxy_results}E) or semi-supervised (Fig. \ref{fig3:proxy_results}C \& \ref{fig3:proxy_results}F). This is also summarized in Table. 1. All methods when inter-compared by validation Dice score per volume are statistically significant (p<<0.05) using Wilcoxon signed rank test.

Qualitatively, the quality of labels generated by semi-supervised models per active iterations are better as compared to supervised models (Ref Fig. \ref{fig4:qual_results}). The uncertainty maps indicate more relevant ROIs which assists in selecting more relevant data. The bottom row shows an intuition of how proxy rank selects data, we can observe that unlabeled volumes with abnormalities (tumor) tend to have a higher uncertainty as compared to volumes without abnormalities.

\section{Conclusion}

Proxy-based ranking and semi-supervised learning both add much needed tools to AL. Where we see typical supervised learning based AL fail or lack in performance, the proposed strategies can add significant leverage. It should be noted that the HU-based proxy ranking is limited to CT datasets; future work could focus on how a proxy rank could be estimated for other imaging modalities such as magnetic resonance imaging, ultra-sound, etc.

\bibliographystyle{splncs04}
\bibliography{main}

\begin{thebibliography}{10}
\providecommand{\url}[1]{\texttt{#1}}
\providecommand{\urlprefix}{URL }
\providecommand{\doi}[1]{https://doi.org/#1}

\bibitem{ash2019deep}
Ash, J.T., Zhang, C., Krishnamurthy, A., Langford, J., Agarwal, A.: Deep batch
  active learning by diverse, uncertain gradient lower bounds. arXiv preprint
  arXiv:1906.03671  (2019)

\bibitem{beluch2018power}
Beluch, W.H., Genewein, T., N{\"u}rnberger, A., K{\"o}hler, J.M.: The power of
  ensembles for active learning in image classification. In: Proceedings of the
  IEEE Conference on Computer Vision and Pattern Recognition. pp. 9368--9377
  (2018)

\bibitem{bengar2021deep}
Bengar, J.Z., Raducanu, B., Weijer, J.v.d.: When deep learners change their
  mind: Learning dynamics for active learning. In: International Conference on
  Computer Analysis of Images and Patterns. pp. 403--413. Springer (2021)

\bibitem{bengar2021reducing}
Bengar, J.Z., van~de Weijer, J., Twardowski, B., Raducanu, B.: Reducing label
  effort: Self-supervised meets active learning. In: Proceedings of the
  IEEE/CVF International Conference on Computer Vision. pp. 1631--1639 (2021)

\bibitem{chitta2018large}
Chitta, K., Alvarez, J.M., Lesnikowski, A.: Large-scale visual active learning
  with deep probabilistic ensembles. arXiv preprint arXiv:1811.03575  (2018)

\bibitem{gal2016dropout}
Gal, Y., Ghahramani, Z.: Dropout as a bayesian approximation: Representing
  model uncertainty in deep learning. In: international conference on machine
  learning. pp. 1050--1059 (2016)

\bibitem{gal2017deep}
Gal, Y., Islam, R., Ghahramani, Z.: Deep bayesian active learning with image
  data. In: Proceedings of the 34th International Conference on Machine
  Learning-Volume 70. pp. 1183--1192. JMLR. org (2017)

\bibitem{he2021dints}
He, Y., Yang, D., Roth, H., Zhao, C., Xu, D.: Dints: Differentiable neural
  network topology search for 3d medical image segmentation. In: Proceedings of
  the IEEE/CVF Conference on Computer Vision and Pattern Recognition. pp.
  5841--5850 (2021)

\bibitem{houlsby2014cold}
Houlsby, N., Hern{\'a}ndez-Lobato, J.M., Ghahramani, Z.: Cold-start active
  learning with robust ordinal matrix factorization. In: International
  Conference on Machine Learning. pp. 766--774. PMLR (2014)

\bibitem{isensee2021nnu}
Isensee, F., Jaeger, P.F., Kohl, S.A., Petersen, J., Maier-Hein, K.H.: nnu-net:
  a self-configuring method for deep learning-based biomedical image
  segmentation. Nature methods  \textbf{18}(2),  203--211 (2021)

\bibitem{kingma2014adam}
Kingma, D.P., Ba, J.: Adam: A method for stochastic optimization. arXiv
  preprint arXiv:1412.6980  (2014)

\bibitem{kuo2018cost}
Kuo, W., H{\"a}ne, C., Yuh, E., Mukherjee, P., Malik, J.: Cost-sensitive active
  learning for intracranial hemorrhage detection. In: International Conference
  on Medical Image Computing and Computer-Assisted Intervention. pp. 715--723.
  Springer (2018)

\bibitem{lai2021joint}
Lai, Z., Wang, C., Oliveira, L.C., Dugger, B.N., Cheung, S.C., Chuah, C.N.:
  Joint semi-supervised and active learning for segmentation of gigapixel
  pathology images with cost-effective labeling. In: Proceedings of the
  IEEE/CVF International Conference on Computer Vision. pp. 591--600 (2021)

\bibitem{li2020shape}
Li, S., Zhang, C., He, X.: Shape-aware semi-supervised 3d semantic segmentation
  for medical images. In: International Conference on Medical Image Computing
  and Computer-Assisted Intervention. pp. 552--561. Springer (2020)

\bibitem{nath2021power}
Nath, V., Yang, D., Hatamizadeh, A., Abidin, A.A., Myronenko, A., Roth, H.R.,
  Xu, D.: The power of proxy data and proxy networks for hyper-parameter
  optimization in medical image segmentation. In: International Conference on
  Medical Image Computing and Computer-Assisted Intervention. pp. 456--465.
  Springer (2021)

\bibitem{nath2020diminishing}
Nath, V., Yang, D., Landman, B.A., Xu, D., Roth, H.R.: Diminishing uncertainty
  within the training pool: Active learning for medical image segmentation.
  IEEE Transactions on Medical Imaging  \textbf{40}(10),  2534--2547 (2020)

\bibitem{nguyen2021measure}
Nguyen, V.L., Shaker, M.H., H{\"u}llermeier, E.: How to measure uncertainty in
  uncertainty sampling for active learning. Machine Learning pp. 1--34 (2021)

\bibitem{ouyang2020self}
Ouyang, C., Biffi, C., Chen, C., Kart, T., Qiu, H., Rueckert, D.:
  Self-supervision with superpixels: Training few-shot medical image
  segmentation without annotation. In: European Conference on Computer Vision.
  pp. 762--780. Springer (2020)

\bibitem{ronneberger2015u}
Ronneberger, O., Fischer, P., Brox, T.: U-net: Convolutional networks for
  biomedical image segmentation. In: International Conference on Medical image
  computing and computer-assisted intervention. pp. 234--241. Springer (2015)

\bibitem{sener2017active}
Sener, O., Savarese, S.: Active learning for convolutional neural networks: A
  core-set approach. arXiv preprint arXiv:1708.00489  (2017)

\bibitem{settles2012active}
Settles, B.: Active learning. Synthesis Lectures on Artificial Intelligence and
  Machine Learning  \textbf{6}(1),  1--114 (2012)

\bibitem{simeoni2021rethinking}
Sim{\'e}oni, O., Budnik, M., Avrithis, Y., Gravier, G.: Rethinking deep active
  learning: Using unlabeled data at model training. In: 2020 25th International
  Conference on Pattern Recognition (ICPR). pp. 1220--1227. IEEE (2021)

\bibitem{simpson2019large}
Simpson, A.L., Antonelli, M., Bakas, S., Bilello, M., Farahani, K.,
  Van~Ginneken, B., Kopp-Schneider, A., Landman, B.A., Litjens, G., Menze, B.,
  et~al.: A large annotated medical image dataset for the development and
  evaluation of segmentation algorithms. arXiv preprint arXiv:1902.09063
  (2019)

\bibitem{sourati2018active}
Sourati, J., Gholipour, A., Dy, J.G., Kurugol, S., Warfield, S.K.: Active deep
  learning with fisher information for patch-wise semantic segmentation. In:
  Deep Learning in Medical Image Analysis and Multimodal Learning for Clinical
  Decision Support, pp. 83--91. Springer (2018)

\bibitem{tang2021self}
Tang, Y., Yang, D., Li, W., Roth, H., Landman, B., Xu, D., Nath, V.,
  Hatamizadeh, A.: Self-supervised pre-training of swin transformers for 3d
  medical image analysis. arXiv preprint arXiv:2111.14791  (2021)

\bibitem{wang2018test}
Wang, G., Li, W., Aertsen, M., Deprest, J., Ourselin, S., Vercauteren, T.:
  Test-time augmentation with uncertainty estimation for deep learning-based
  medical image segmentation  (2018)

\bibitem{wang2020semi}
Wang, J., Wen, S., Chen, K., Yu, J., Zhou, X., Gao, P., Li, C., Xie, G.:
  Semi-supervised active learning for instance segmentation via scoring
  predictions. arXiv preprint arXiv:2012.04829  (2020)

\bibitem{wang2021annotation}
Wang, S., Li, C., Wang, R., Liu, Z., Wang, M., Tan, H., Wu, Y., Liu, X., Sun,
  H., Yang, R., et~al.: Annotation-efficient deep learning for automatic
  medical image segmentation. Nature communications  \textbf{12}(1),  1--13
  (2021)

\bibitem{xia20203d}
Xia, Y., Liu, F., Yang, D., Cai, J., Yu, L., Zhu, Z., Xu, D., Yuille, A., Roth,
  H.: 3d semi-supervised learning with uncertainty-aware multi-view
  co-training. In: Proceedings of the IEEE/CVF Winter Conference on
  Applications of Computer Vision. pp. 3646--3655 (2020)

\bibitem{yang2017suggestive}
Yang, L., Zhang, Y., Chen, J., Zhang, S., Chen, D.Z.: Suggestive annotation: A
  deep active learning framework for biomedical image segmentation. In:
  International conference on medical image computing and computer-assisted
  intervention. pp. 399--407. Springer (2017)

\bibitem{yu2019uncertainty}
Yu, L., Wang, S., Li, X., Fu, C.W., Heng, P.A.: Uncertainty-aware
  self-ensembling model for semi-supervised 3d left atrium segmentation. In:
  International Conference on Medical Image Computing and Computer-Assisted
  Intervention. pp. 605--613. Springer (2019)

\bibitem{yuan2020cold}
Yuan, M., Lin, H.T., Boyd-Graber, J.: Cold-start active learning through
  self-supervised language modeling. arXiv preprint arXiv:2010.09535  (2020)

\end{thebibliography}

%
%
%
%
%



\end{document}


\title{Supplementary Material: Warm Start Active Learning with Proxy Labels \& Selection via Semi-Supervised Fine-Tuning}
\titlerunning{Proxy Semi-Supervised Active Learning}
%
\author{Vishwesh Nath\inst{} \and
Dong Yang\inst{} \and
Holger R. Roth\inst{} \and
Daguang Xu\inst{}}
%
\authorrunning{V. Nath et al.}
%
\institute{NVIDIA
\email{}\\
\url{}
\email{vnath@nvidia.com}}
\maketitle              
\section{Additional Experiment on Hepatic Vessel \& Tumor Dataset}

In the main paper, the experiment results shared for hepatic vessels \& tumor dataset \cite{simpson2019large} is with a limit of 4 active iterations. This was done so to keep the experimental settings consistent with the other dataset of liver \& tumor \cite{simpson2019large}. 

\begin{figure}[!h]
\begin{center}
    \includegraphics[width=0.5\linewidth]{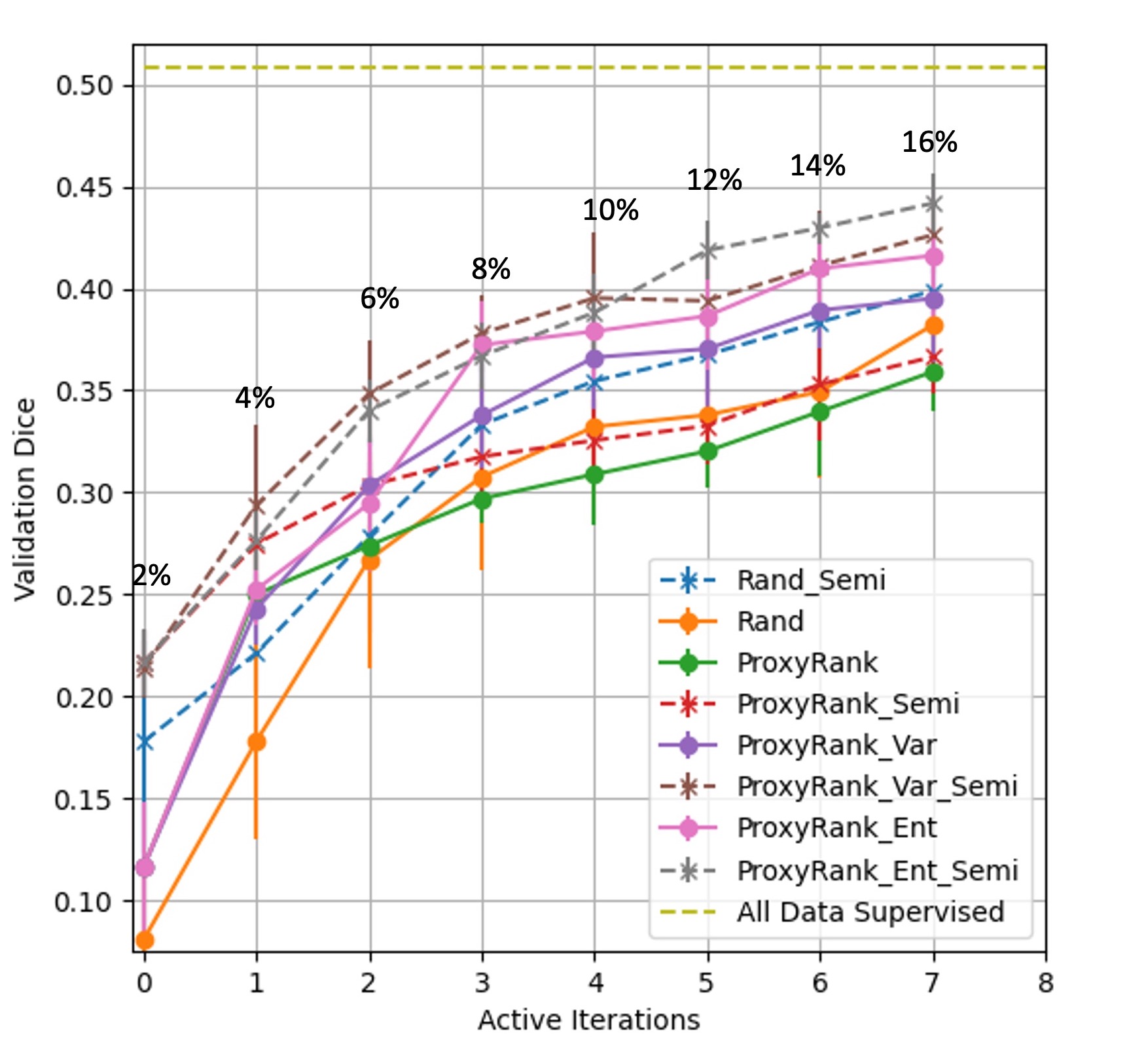}
    \caption{The percentage represents the amount of data used at the current active iteration with respect to all available data. Solid lines represent fully supervised approaches and dashed lines represent semi-supervised approaches. ``Proxyrank'' indicates that the initial training pool was estimated using proxy task. ``Var'' $\rightarrow$ ``Variance'', ``Ent'' $\rightarrow$ ``Entropy'', ``Rand'' $\rightarrow$ ``Random'', ``Semi'' $\rightarrow$ ``Semi-supervised learning''}
    \label{supp_fig:hepatic}
\end{center}
\end{figure}

The additional experiment performed is on the hepatic vessel \& tumor dataset for 8 active iterations (Fig. S\ref{supp_fig:hepatic}) instead of 4 as in the main paper. The result of 4 active iterations can be observed in Fig. \textcolor{red}{2} in the main paper. The other experimental hyper-parameters for this experiment are consistent with the ones used in the main paper as outlined in section \textcolor{red}{4.2}. The motivation being that the active learning methods when using only $8\%$ of data as labeled data still have a significant gap with the model that is using all the data. 

With more active iterations, the amount of labeled data that are being used increases from $8\%$ to $16\%$ and the active learning models are much closer to the model using all data. It can also be observed that the semi-supervised active learning methods along with initial selection of data via proxy ranking provide a significant benefit over the supervised active learning methods.

\begin{table*}[!t]
  \centering
  \scriptsize
  \begin{tabular}{p{3.0cm} p{2.6cm} p{1.5cm} p{1.5cm} p{1.5cm} p{1.5cm}  }
     \hline
     \multicolumn{6}{c}{\textbf{Proxy Rank \& Semi-Supervised Active Learning Ablation Settings}} \\
     \hline
     \textit{Setting} & Plot Name & Proxy Ranking & Pre-trained Weights & Semi-Supervised & Acquisition Function  \\\hline
     Random & Rand & \xmark & \xmark & \xmark & \xmark \\
     Variance & Var & \xmark & \xmark & \xmark & \cmark \\
     Entropy & Ent & \xmark & \xmark & \xmark & \cmark \\
     Variance SS & Var\textunderscore Semi & \xmark & \xmark & \cmark & \cmark \\
     Entropy SS & Ent\textunderscore Semi & \xmark & \xmark & \cmark & \cmark \\
     Random SS & Rand\textunderscore Semi & \xmark & \xmark & \cmark & \xmark \\
     Proxy Rank & ProxyRank & \cmark & \cmark & \xmark & \xmark\\
     Proxy Rank SS & ProxyRank\textunderscore Semi & \cmark & \cmark & \cmark & \xmark \\
     Proxy Rank Variance & ProxyRank\textunderscore Var & \cmark & \cmark & \xmark & \cmark\\
     Proxy Rank Variance SS & ProxyRank\textunderscore Var\textunderscore Semi& \cmark & \cmark & \cmark & \cmark\\
     Proxy Rank Entropy & ProxyRank\textunderscore Ent & \cmark & \cmark & \xmark & \cmark\\
     Proxy Rank Entropy SS & ProxyRank\textunderscore Ent\textunderscore Semi& \cmark & \cmark & \cmark & \cmark\\
     
  \end{tabular}
  \caption{Different experimental settings that were tested as ablations for the experiments. SS $\rightarrow$ Semi-Supervised. The associated plot legend name is also given for each method for both Figure 2 and 3 from the paper}
  \label{tab:method_select}
\end{table*}

\bibliographystyle{splncs04}
\bibliography{egbib}

%
%
%
%
%


